\documentclass[11pt]{article}

\usepackage[final]{ACL2023}

\usepackage{times}
\usepackage{latexsym}
\usepackage{graphicx} 
\usepackage{booktabs}
\usepackage{tabularx}
\usepackage[utf8]{inputenc}
\usepackage{amsmath}

\usepackage[T1]{fontenc}
\usepackage[utf8]{inputenc}
\usepackage{booktabs}

\usepackage{microtype}

\usepackage{inconsolata}

\title{RoundTripOCR: A Data Generation Technique for Enhancing Post-OCR Error Correction in Low-Resource Devanagari Languages}

\author{Harshvivek Kashid  \and Pushpak Bhattacharyya \\
  Indian Institute of Technology Bombay \\
  \texttt{\{harshvivek,pb\}@cse.iitb.ac.in}\\}

\begin{document}
\maketitle
% TODO
% \begin{itemize}
%     \item Add - real test section (results)
%     \item Qualitative/Error Analysis

% \end{itemize}

\begin{abstract}
Optical Character Recognition (OCR) technology has revolutionized the digitization of printed text, enabling efficient data extraction and analysis across various domains. Just like Machine Translation systems, OCR systems are prone to errors. In this work, we address the challenge of data generation and post-OCR error correction, specifically for low-resource languages. We propose an approach for synthetic data generation for Devanagari languages, \textbf{\textit{RoundTripOCR}}, that tackles the scarcity of the post-OCR Error Correction datasets for low-resource languages. We release post-OCR text correction datasets for Hindi, Marathi, Bodo, Nepali, Konkani and Sanskrit. We also present a novel approach for OCR error correction by leveraging techniques from machine translation. Our method involves translating erroneous OCR output into a corrected form by treating the OCR errors as mistranslations in a parallel text corpus, employing pre-trained transformer models to learn the mapping from erroneous to correct text pairs, effectively correcting OCR errors.
\end{abstract}

\begin{figure}[ht]
    \centering
\includegraphics[width=1\columnwidth]
    {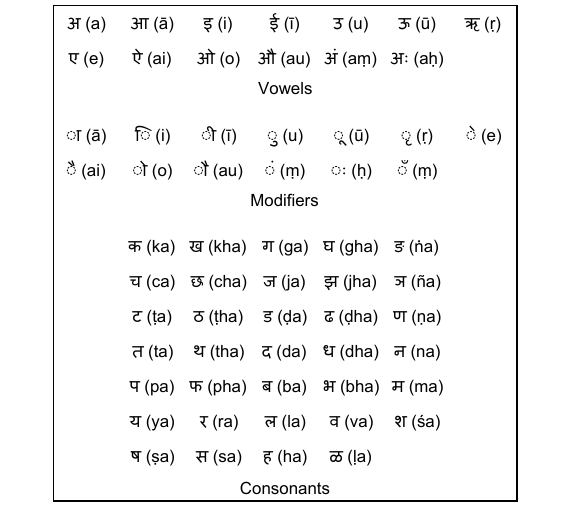}
\caption{ Vowels, modifiers and consonants of Devanagari script.} 
    \label{fig:chars}
\end{figure}

\section{Introduction}
The Devanagari script is the most extensively used writing system in the Indian subcontinent. It was the principal script for Sanskrit, the ancient literary language of Indian civilization. Sanskrit was used to write a wide range of texts covering various domains, including literature, philosophy, science, art, architecture, and mathematics. This includes the Vedas, Upanishads, and epics like Mahabharata and Ramayana. Devanagari script originated from ancient Brahmi script through various transformations \cite{dev}. Apart from vowels, modifiers and consonants (Figure \ref{fig:chars}), it has a rich set of conjunct consonants, known as ligatures, where multiple characters combine to form new glyphs. These are difficult to segment and recognize because they don't correspond directly to individual letters. Devanagari characters often have vowel signs (\textit{matras}) and other diacritical marks that appear above, below, or beside the base character. These modifiers can be challenging to detect, segment, and associate correctly with the base character. Devanagari script includes a horizontal line (called the \textit{Shirorekha}) that connects the characters in each word. Unlike in Latin scripts, where spaces clearly divide words, in Devanagari, words often connect via the headline, making word segmentation harder for OCR systems.
 Proper segmentation of Devanagari words, characters, and sub-components (such as vowels and consonants) is difficult because components often overlap, connect through ligatures, or blend with the \textit{Shirorekha} line. This is less common in simpler scripts like Latin, where individual letters are often spaced apart and stand independently. Many Devanagari characters look quite similar, especially in certain fonts or degraded images, leading to higher chances of OCR errors.
OCR technology has revolutionized the digitization and processing of written or printed text by enabling machines to automatically convert scanned documents or handwritten texts into editable and searchable text formats. However, despite significant advancements over the years, the accurate recognition of text from scanned documents remains a challenging task due to inherent complexities in document layouts, font variations, noise, and other distortions.

Traditional OCR systems typically follow a pipeline approach comprising image preprocessing, feature extraction, character segmentation, and recognition stages. While these systems have achieved remarkable success in many applications, they are susceptible to errors, especially when dealing with degraded or low-quality document images. OCR errors can manifest in various forms, including misrecognitions, substitutions, omissions, and insertions, leading to inaccuracies in the recognized text output. These errors not only impede the reliability of OCR systems but also pose significant challenges for downstream tasks such as information extraction, text mining, and machine translation \cite{kolak-etal-2003-generative, ocr-appli, survey_ocr, ignat-etal-2022-ocr}. Addressing OCR errors requires robust error detection and correction mechanisms that can effectively handle a wide range of error patterns and variations.

\textbf{Our contributions are:}
\begin{enumerate}
    \item \textbf{RoundTripOCR}\footnote{RoundTripOCR code and dataset details are on GitHub: \url{https://github.com/harshvivek14/RoundTripOCR}}, a technique to artificially generate post-OCR error correction data for low-resource Devanagari script languages in the form <\textit{T, T\textbf{'}}>, where \textit{T\textbf{'}} is the OCR output text and \textit{T} is the correct OCR output text (Section \ref{sec:roundtrip}).
    
    \item Post-OCR error correction dataset, containing 3.1 million sentences in Hindi, 1.58 million sentences in Marathi, 2.54 million sentences in Bodo, 2.97 million sentences in Nepali, 1.95 million sentences in Konkani and 4.07 million sentences in Sanskrit (Table \ref{tab:dataset_distribution}).
    
    \item Benchmarks for the Post-OCR error correction task based on the pre-trained Seq2Seq language models for all six languages (Section \ref{sec:results}).

\end{enumerate}
\section{Related work}
%  The goal of OCR error correction is to transform the erroneous OCR output into a text that matches the ground truth. Given an OCR-generated sequence $X = \{x_1, x_2, \dots, x_n\}$ and the corresponding ground truth sequence $Y = \{y_1, y_2, \dots, y_m\}$, the objective is to find a corrected sequence $\hat{Y}$ that closely matches $Y$.

% Formally, the goal is to find $\hat{Y}$ that minimizes the difference between $\hat{Y}$ and $Y$, where the difference is often measured in terms of character or word-level discrepancies:

% \[
% \hat{Y} = \arg\min_Y \text{Difference}(X, Y)
% \]

% Here, $\text{Difference}(X, Y)$ represents the error metric used to quantify discrepancies between the corrected sequence and the ground truth. Addressing OCR errors is crucial for improving the overall quality and usability of digitized documents.

As mentioned by \citet{ocrerror1} and \citet{ocrerror2}, OCR systems are prone to various types of errors that can occur during the process of text recognition from scanned documents. 
The most common types of OCR errors include:
substitution errors, insertion errors, deletion errors and segmentation errors.

Even state-of-the-art OCR models are susceptible to making recognition errors \citep{dong-smith-2018-multi}. Errors are particularly frequent in the case of low-resource languages because most off-the-shelf OCR tools do not directly support these languages, and training a high-performance OCR system is challenging given the small amount of data that is typically available \citep{rijhwani-etal-2020-ocr}. We use post-OCR error correction tools and techniques to correct these errors and improve the quality of the transcription.
Over the years, researchers have explored various approaches to mitigate OCR errors, including rule-based post-processing techniques \citep{ruleocr}, statistical language models \citep{statocr}, and machine learning-based methods \citep{virk-etal-2021-novel}. While these approaches have shown promise in certain scenarios, they often rely on handcrafted rules or linguistic resources, limiting their generalization to diverse document types and languages.

In recent years, there has been growing interest in applying advanced machine learning and natural language processing techniques to address OCR errors effectively. One promising direction is to leverage techniques from machine translation, which aims to automatically translate text from one language to another \citep{lyu-etal-2021-neural}. By treating OCR errors as mistranslations and modelling the correction process as an automatic post-editing (APE) task, it is possible to harness the power of neural machine translation models to learn the mapping from erroneous to correct OCR text output. This paradigm shift not only enables end-to-end error correction but also facilitates the integration of contextual information and linguistic knowledge into the correction process, leading to more accurate and robust OCR systems. 

The emergence of Transformer architecture and attention mechanisms \cite{vaswani2023attentionneed} has led to the adoption of deep learning models for post-OCR tasks. Post-OCR tasks have been reframed as Sequence-to-Sequence tasks in recent studies, whereby researchers have applied Machine Translation models \cite{nmt-smt-ocr}. The BERT \cite{devlin-etal-2019-bert} and BART \cite{lewis-etal-2020-bart} models were used by \citet{nguyen2020} and \citet{soper-etal-2021-bart}, respectively. \citet{maheshwari-etal-2022-benchmark} compared standard Sequence-to-Sequence models with pre-trained models. A lot of data is needed to train these models, and the predominant method for obtaining post-OCR training data has been crowdsourcing \cite{clematide-etal-2016-crowdsourcing}. Although this can yield extremely accurate training data, the procedure often proves costly and time-consuming. Thus, synthetic data generation has been widely employed in this task \cite{dhondt-etal-2017-generating, jasonarson-etal-2023-generating, guan-greene-2024-advancing}. The sentence or line-level OCR error correction by using the sentence or line-level dataset has also proven to be effective in addressing segmentation and word errors of the OCR output \cite{line-level-ocr1, lyu-etal-2021-neural, rijhwani-etal-2021-lexically, ignat-etal-2022-ocr, thomas-etal-2024-leveraging}.

\subsection{Automatic Post-Editing and OCR Error Correction }
Automatic Post-Editing (APE) uses techniques to improve the quality of Machine Translation (MT) output automatically, including rule-based, statistical, and neural-based techniques \citep{chollampatt-etal-2020-automatic, deoghare-etal-2023-quality}. APE systems are trained on human-edited translations, allowing them to identify and correct errors in grammar, fluency, and terminology. While MT systems have advanced significantly, they often produce translations that contain errors or lack fluency, especially with complex or domain-specific content.  Output generated by a machine translation system is not always perfect and hence requires further editing \citep{parton-etal-2012-automatic,laubli-etal-2013-assessing,pal-etal-2016-multi}. 

OCR systems play a crucial role in digitizing text, but inherent limitations lead to errors in the extracted text. This necessitates post-processing techniques to refine the OCR output and achieve higher accuracy \citep{ocrappproaches}. Viewing this process through the lens of APE offers a valuable framework for developing effective error correction methods. Post-OCR error correction can be considered an Automatic Post Editing task. Similar to a machine translation system generating a translated sentence from a source language, the OCR system produces a text present in an image. This process is prone to errors due to limitations in OCR systems, image quality, and stylistic variation. Just like an APE system refines a machine-translated sentence to improve fluency and accuracy, the post-OCR correction system aims to refine the text generated by the OCR system to remove errors and achieve a more accurate representation of the original document. Both MT and OCR error correction face common challenges like handling ambiguity, dealing with rare words, and adapting to stylistic variations.

\subsection{Round-trip translation}
Synthetic data generation techniques are generally employed to generate artificial data for training machine learning models and neural networks. Due to insufficient post-editing data available for the WMT APE 2016 shared task \cite{wmt2016} to train neural models, \citet{junczys2016log}, created two phrase-based translation models: 
English-German and German-English, using provided parallel training data to conduct round-trip translation. Using them in the Round-trip Translation approach resulted in the generation of artificial post-editing triplets \textit{<src, mt, pe>}, where \textit{src} is source sentence, \textit{mt} is machine translated sentence and \textit{pe} is post-edited sentence.
This artificial data creation method assisted in resolving the problem of insufficient training data, which frequently arises in NMT-based systems. Inspired by the Round-trip Translation approach and image-based synthetic data generation technique for the OCR system by \citet{recipe}, which promises unlimited training data at zero annotation cost, we propose a synthetic data generation technique for post-OCR error correction, \textbf{RoundTripOCR},  which we discuss in detail in the following section.

\begin{table*}[h]
    \centering
    \begin{tabular}{|l|r|r|r|r|r|r|}
        \hline
        \# of Sent. & \textbf{Hindi} & \textbf{Marathi} & \textbf{Bodo} & \textbf{Nepali} & \textbf{Konkani} & \textbf{Sanskrit} \\
        \hline
        Train dataset & 3,129,200 & 1,581,405 & 2,541,649 & 2,970,148 & 1,950,874 & 4,070,000
 \\
        % \hline
        Validation set & 10,000 & 10,000 & 10,000 & 10,000 & 10,000 & 10,000 \\
        % \hline
        Test dataset & 10,000 & 10,000 & 10,000 & 10,000 & 10,000 & 10,000 \\
        \hline
    \end{tabular}
    \caption{Distribution of dataset generated using RoundTripOCR technique for all six languages.}
    \label{tab:dataset_distribution}
\end{table*}

\begin{figure}[h] % h means place the figure here if possible, or t for top, b for bottom, etc.
\centering
\includegraphics[width=0.7\columnwidth]{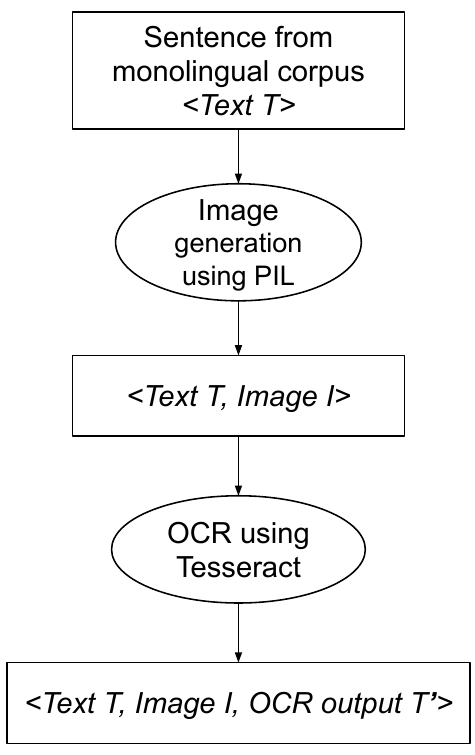}
\caption{\textbf{RoundTripOCR}: Artificial post-OCR error correction data generation process. We get  <\textit{Text T, Image I, OCR output T\textbf{'}}>  as output, where <\textit{Text T}> will be used as corrected OCR output text and <\textit{OCR output T\textbf{'}}> as OCR output.}
\label{fig:ocr}
\end{figure}

\section{RoundTripOCR}\label{sec:roundtrip}

\begin{figure*}[ht]
    \centering
    \includegraphics[width=2\columnwidth]
    {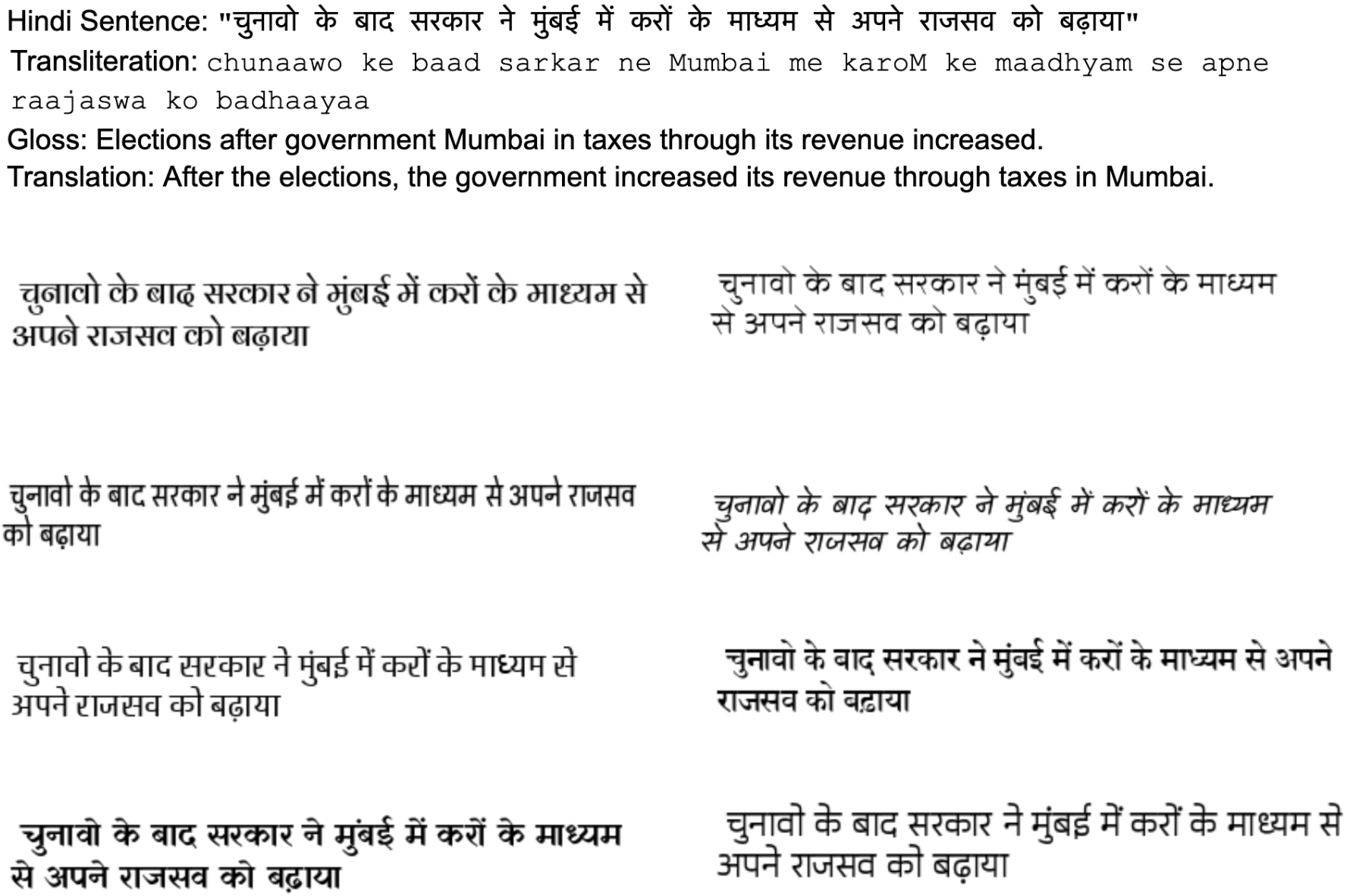}
    \hspace{1mm}
    \caption{Examples of images generated with different fonts during RoundTripOCR data generation process.} 
    \label{fig:fonts}
\end{figure*}

The creation of artificial OCR data involves a systematic process aimed at simulating real-world scenarios while taking into consideration the common OCR error types and generating diverse datasets for training and evaluation purposes.

To introduce variability into the dataset, 50 different Devanagari font combinations were selected from Google Fonts\footnote{\url{https://fonts.google.com/?subset=devanagari}}. Each font style offered unique characteristics, such as varying stroke thickness, serif styles, and overall aesthetics, as shown in Figure \ref{fig:fonts}. Utilizing the selected Devanagari font combinations, 50 images could potentially be generated from a single sentence. PIL provides a comprehensive set of image processing functionalities, enabling the programmatic creation of images with text rendered in specific font styles. The generated images were subjected to optical character recognition (OCR) using the Pytesseract library. Pytesseract is not supported for Bodo, Nepali, and Konkani languages. Thus, we use Pytesseract-Hindi for Bodo and Nepali along with Hindi and Pytesseract-Marathi for Konkani and Marathi due to similarities in these languages. We used Pytesseract-Sanskrit for the Sanskrit language. Pytesseract leverages machine-learning algorithms to extract text from images and convert them into machine-readable formats, including the Devanagari texts. The OCR process is aimed at simulating real-world OCR scenarios and generating text outputs from the rendered images. Since we can get 50 \textit{<Text T, OCR output T\textbf{'}>} data points from a single sentence \textit{<Text T>}, this approach can be extended to any low-resource language.

By following this methodology, as shown in Figure \ref{fig:ocr}, a comprehensive artificial dataset for OCR error correction was generated, encompassing a diverse range of text passages, font styles, and linguistic variations. This dataset serves as a valuable resource for training and evaluating OCR systems, enabling researchers and practitioners to develop robust OCR algorithms and assess their performance under various conditions.

\subsection{Dataset}
\label{sec:dataset}
We generate post-OCR error correction datasets for Bodo, Nepali, Konkani, Hindi, Sanskrit and Marathi texts. The corpora for Hindi was sourced from the CC-100 corpus \cite{conneau-etal-2020-unsupervised}, and Konkani, Nepali, Bodo and Marathi texts were sourced from Technology Development for Indian Languages (TDIL)\footnote{\url{https://www.tdil-dc.in}} and Sanskrit texts were sourced from  \citet{maheshwari-etal-2022-benchmark}.
Leveraging the RoundTripOCR technique, we generate datasets containing around 3.1 million sentence pairs in Hindi, 1.58 million sentence pairs in Marathi, 2.54 million sentence pairs in Bodo, 2.97 million sentence pairs in Nepali, 1.95 million sentence pairs in Konkani and 4.07 million sentence pairs in Sanskrit as mentioned in the Table \ref{tab:dataset_distribution}.  Each pair have <\textit{Text T}>, which is the corrected OCR output text, and <\textit{OCR output T\textbf{'}}>, which is the OCR output sentence\footnote{Datasets are available at: \url{https://github.com/harshvivek14/RoundTripOCR}}.

% Font analysis revealed significant variations in error rates. Specifically, fonts such as Khand-Regular, Rajdhani-Regular, Nirmala, and Biryani exhibited the highest CERs, exceeding 7\%. Conversely, fonts like Gargi, Karma-Regular, NotoSans-Regular, and VesperLibre-Regular demonstrated remarkably low CERs, each falling below 1\% as shown in Figure \ref{fig:graph}. Our findings suggest that models trained on a diverse range of fonts perform more robustly than those trained solely on a single font. This observation underscores the importance of font diversity in enhancing OCR error correction models' performance and resilience.

\section{Sequence to Sequence models}

We conducted a series of experiments employing sequence-to-sequence models: \textbf{mBART, mT5} and \textbf{IndicBART}. These are powerful models designed for multilingual tasks, particularly in low-resource languages. 

\textbf{mBART} (Multilingual BART) is a sequence-to-sequence denoising autoencoder that pre-trains on a variety of languages by corrupting and reconstructing text, making it highly effective for tasks like machine translation and text generation across different languages \cite{liu-etal-2020-multilingual-denoising}. It has been extensively used for tasks involving noisy inputs, such as post-OCR error correction, due to its ability to learn contextual representations and perform cross-lingual transfer \cite{soper-etal-2021-bart, maheshwari-etal-2022-benchmark}. We used \textit{mbart-large-50} version of mBART.

\textbf{mT5} (Multilingual T5) extends the T5 model's text-to-text framework to a massively multilingual setting \cite{xue-etal-2021-mt5}. With the capacity to handle over 100 languages, mT5 is effective for multilingual NLP tasks, including translation, summarization, and post-OCR error correction \cite{mt5-ocr}. This model leverages the original T5 framework, where every NLP task is reframed as a text generation problem, allowing for consistent and flexible handling of a wide range of tasks across languages. The version we use in our experiments is \textit{mT5-base}.

\textbf{IndicBART} is a variant of mBART that is specifically tailored for Indic languages like Hindi, Bengali, Marathi, and others \cite{dabre-etal-2022-indicbart}. It adapts the pre-training and fine-tuning processes to better handle the linguistic and scriptural characteristics of these languages, which are often underrepresented in large-scale language models. IndicBART has proven to be highly effective for tasks such as machine translation in Indic scripts.

\section{Experiments and Results}
The pre-trained models were sourced from Hugging Face\footnote{\url{https://huggingface.co/models}} and finetuned using NVIDIA A100 GPU for 2 to 3 epochs. A learning rate of 5e-4 was applied, managed by a polynomial learning rate scheduler. The training was conducted with 32-bit floating-point precision, and the best-performing model from each run was saved for evaluation.
To facilitate effective model training and evaluation, we partitioned the dataset into training, testing, and validation sets. The testing set and validation set contained 10,000 pairs each.

\begin{table*}[h]
\centering
\small % Reduce font size
\resizebox{\textwidth}{!}{
\begin{tabular}{
  c  % Model
  *{2}{p{0.8cm}}  % Hindi CER & WER
  *{2}{p{0.8cm}}  % Marathi CER & WER
  *{2}{p{0.8cm}}  % Konkani CER & WER
  *{2}{p{0.8cm}}  % Nepali CER & WER
  *{2}{p{0.8cm}}  % Bodo CER & WER
  *{2}{p{0.8cm}}  % Sanskrit CER & WER
}
% \hline
 & \multicolumn{2}{c}{\textbf{Hindi}} & \multicolumn{2}{c}{\textbf{Marathi}} & \multicolumn{2}{c}{\textbf{Konkani}} & \multicolumn{2}{c}{\textbf{Nepali}} & \multicolumn{2}{c}{\textbf{Bodo}} & \multicolumn{2}{c}{\textbf{Sanskrit}} \\
 \cmidrule(lr){2-3} \cmidrule(lr){4-5} \cmidrule(lr){6-7} \cmidrule(lr){8-9} \cmidrule(lr){10-11} \cmidrule(lr){12-13}
\textbf{Model} & \textbf{CER} & \textbf{WER} & \textbf{CER} & \textbf{WER} & \textbf{CER} & \textbf{WER} & \textbf{CER} & \textbf{WER} & \textbf{CER} & \textbf{WER} & \textbf{CER} & \textbf{WER} \\
\hline
OCR (\textit{Tesseract}) & 2.25\% & 5.83\% & 4.10\% & 15.37\% & 4.22\% & 16.80\% & 5.78\% & 24.29\% & 5.89\% & 24.03\% & 8.77\% & 44.73\% \\
IndicBART (single) & 2.30\% & 5.65\% & 4.23\% & 15.04\% & 3.78\% & 14.64\% & 5.56\% & 22.94\% & 4.69\% & 16.69\% & 6.84\% & 32.31\% \\
IndicBART (\textit{all fonts}) & 2.19\% & 5.33\% & 4.08\% & 12.95\% & 3.51\% & 12.70\% & 4.04\% & 15.04\% & 3.84\% & 12.65\% & 6.38\% & 31.89\% \\
mT5 (\textit{single}) & 2.08\% & 5.50\% & 3.65\% & 15.01\% & 3.81\% & 15.38\% & 3.88\% & 16.51\% & 4.45\% & 16.99\% & 6.57\% & 30.23\% \\
mT5 (\textit{all fonts}) & 1.95\% & 4.88\% & 2.91\% & 10.51\% & 3.13\% & 12.95\% & 3.37\% & 14.29\% & 4.20\% & 15.53\% & 6.41\% & 29.32\% \\
mBART (\textit{single}) & 2.11\% & 5.82\% & 3.59\% & 14.47\% & 3.28\% & 13.06\% & 3.19\% & 14.27\% & 3.68\% & 13.02\% & 6.43\% & 29.29\% \\
mBART (all fonts) & \textbf{1.56\%} & \textbf{3.47\%} & \textbf{2.46\%} & \textbf{9.89\%} & \textbf{2.27\%} & \textbf{8.52\%} & \textbf{2.39\%} & \textbf{10.65\%} & \textbf{2.36\%} & \textbf{6.82\%} & \textbf{5.67\%} & \textbf{25.50\%} \\
\hline
\end{tabular}}
\caption{Comparison of mBART, mT5, and IndicBART for Hindi, Marathi, Konkani, Nepali, Bodo, and Sanskrit test datasets based on CER and WER metrics. Tesseract OCR is the baseline. Models for which training is done using a single font style data are indicated as: \textit{single}. Models trained on data with all fonts are indicated as \textit{all fonts}. The best results are highlighted in bold. }
\label{tab:ocr_comparison}
\end{table*}

\begin{figure*}[h]
    \centering
    \includegraphics[width=2\columnwidth]
    {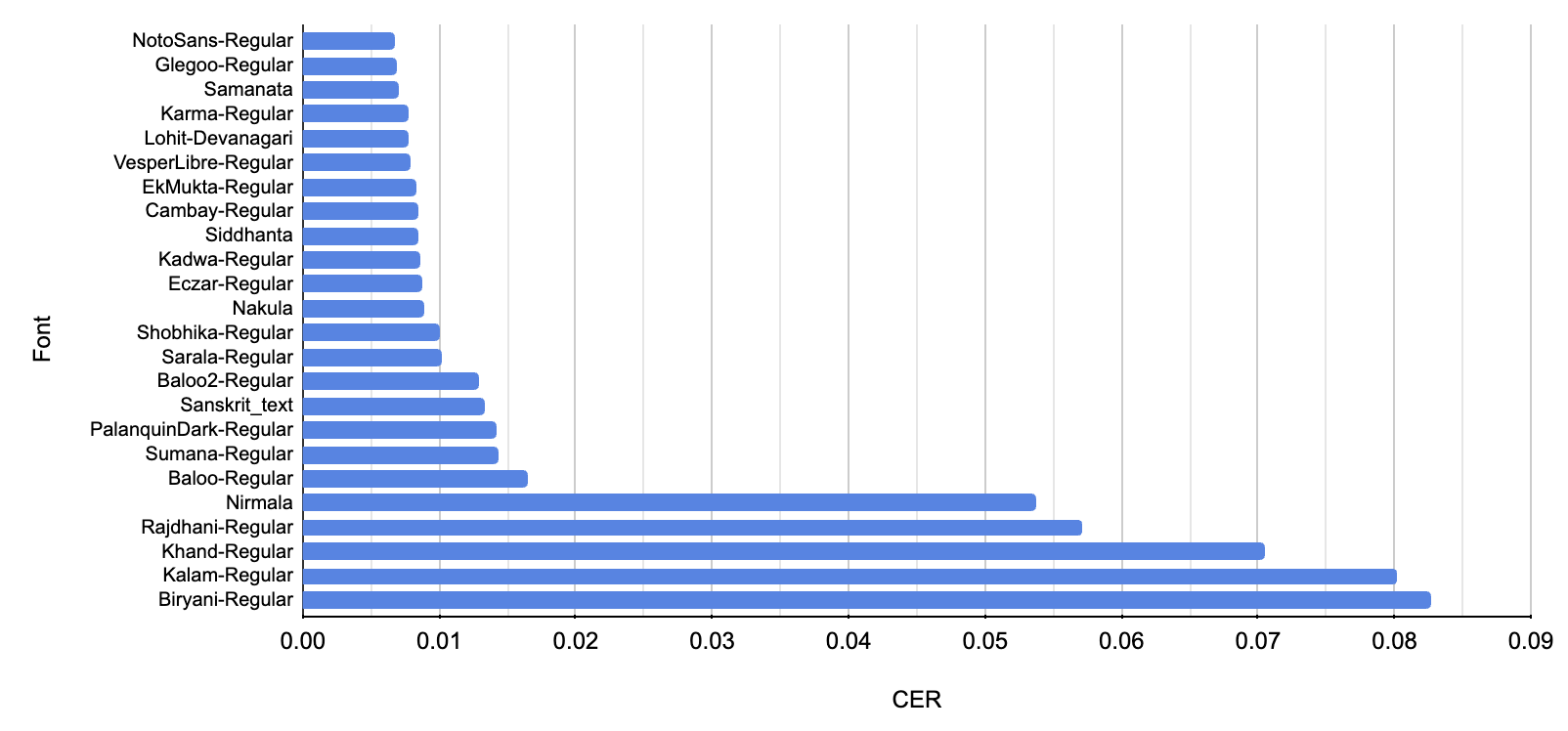}
    \hspace{1mm}
    \caption{Comparision of different fonts and their CER in the Hindi test dataset.} 
    \label{fig:graph}
\end{figure*}

We further curated the second dataset exclusively featuring a single font style; in particular, we chose the \textit{Sumana} font as it shows a close to average CER when compared with all the fonts used in the creation of the dataset as shown in Figure \ref{fig:graph}. This bifurcation allowed us to explore the potential advantages conferred by employing data with varying font styles, thereby enriching our understanding of the model's performance under different font conditions. 

\subsection{Evaluation metric}

In OCR error correction, performance is commonly measured using \textit{Character Error Rate} (CER) and \textit{Word Error Rate} (WER). Both metrics evaluate the edit distance between predicted and ground truth text.

\textbf{CER} is defined as:

\[
CER = \frac{Sc + Dc + Ic}{N}
\]

where $Sc$, $Dc$, and $Ic$ are the number of character-level substitutions, deletions, and insertions, respectively, and $N$ is the total number of characters in the reference text.

\textbf{WER} is similarly defined at the word level:

\[
WER = \frac{Sw + Dw + Iw}{W}
\]

$Sw$, $Dw$, and $Iw$ are the number of word-level substitutions, deletions, and insertions, respectively, and $W$ is the total number of words in the reference text. Lower CER and WER indicate better OCR error correction performance.

\subsection{Results}
\label{sec:results}
We evaluated the performance of several models, IndicBART, mT5, and mBART, on six languages: Hindi, Marathi, Konkani, Nepali, Bodo, and Sanskrit. The models were assessed using two metrics: CER and WER. Tesseract output was considered as the baseline. Across all languages, mBART (all fonts) consistently outperformed other models, showing the lowest CER and WER, followed by mT5 (all fonts). We present detailed results of all conducted experiments in Table \ref{tab:ocr_comparison} comparing the finetuned models with the baseline in the test dataset. 

For instance, in Hindi, Tesseract recorded a CER of 2.25\% and a WER of 5.83\%, whereas the neural models significantly reduced the errors. Among them, \textit{mBART (all fonts)} consistently demonstrated the best performance with a CER of 1.56\% and a WER of 3.47\%. Similar trends were observed in Marathi, where Tesseract had a CER of 4.10\% and a WER of 15.37\%, while \textit{mBART (all fonts)} outperformed with a CER of 2.46\% and WER of 9.89\%.

In Konkani, Tesseract's error rates were even higher, with a CER of 4.22\% and a WER of 16.80\%. However, \textit{mBART (all fonts)} again achieved the best results with a CER of 2.27\% and a WER of 8.52\%, illustrating its robust performance across different scripts. Nepali, being another challenging language for OCR, saw a high error rate from Tesseract (CER of 5.78\% and WER of 24.29\%), but \textit{mBART (all fonts)} reduced these errors to a CER of 2.39\% and WER of 10.65\%. For Bodo, Tesseract recorded a CER of 5.89\% and WER of 24.03\%, while \textit{mBART (all fonts)} again provided substantial improvements, bringing the CER down to 2.36\% and the WER to 6.82\%.

Sanskrit presented the greatest challenge, with Tesseract yielding high error rates of 8.77\% CER and 44.73\% WER. Even here, \textit{mBART (all fonts)} outperformed the other models with a CER of 5.67\% and a WER of 25.50\%, marking a significant improvement. We also tested our best-performing model on 1,000 randomly selected unseen sentences from \citet{maheshwari-etal-2022-benchmark}, which were obtained by OCRing Sanskrit books using Tesseract, resulting in a 6.34\% CER and 41.8\% WER. After OCR error correction, we achieved a 3.42\% CER and 25.7\% WER. This improvement in error rates confirms the efficacy of our proposed RoundTripOCR technique in real-world use cases as well. In summary, \textit{mBART (all fonts)} consistently delivered the best results across all languages, reducing both CER and WER considerably compared to raw OCR output from Tesseract, followed closely by \textit{mT5 (all fonts)}. These findings highlight the advantage of transformer-based models for OCR error correction.

\section{Conclusion and Future Work}
We introduced a novel approach for OCR error correction data generation and created a vast dataset comprising 3.1 million sentences in Hindi, 1.58 million sentences in Marathi, 2.54 million sentences in Bodo, 2.97 million sentences in Nepali, 1.95 million sentences in Konkani, and 4.07 million sentences in Sanskrit. Our proposed methodology is versatile and can be extended to other low-resource languages that follow the Devanagari script. By leveraging monolingual corpora, our approach enables the generation of OCR correction datasets, thus addressing the scarcity of data in such languages. 

The findings from our experimentation underscore the efficacy of approaches from Machine Translation for the task of OCR error output correction, specifically state-of-the-art models like \textit{mBART}, trained on diverse datasets to substantially enhance OCR accuracy. Our research contributes to making textual content more accessible and usable, thereby facilitating broader access to information and knowledge in multilingual societies. Our findings also confirm that models trained on a diverse range of fonts perform more robustly than those trained solely on a single font. This observation underscores the importance of font diversity in enhancing OCR error correction models' performance and resilience.

Our findings motivate the exploration of data augmentation techniques utilizing synthetically generated images as future work. By incorporating these images with controlled variations in font styles, noise levels, and image degradations using a synthetic data generator tool\footnote{\url{https://pypi.org/project/trdg}} for text recognition, we can investigate the impact on model generalization and robustness towards real-world document image complexities. We propose the experimental findings in this work as a baseline, based on which future work can focus on novel and sophisticated techniques for the task of OCR error correction and detection, including improvements to the architecture.

\section*{Limitations}
Our work focuses on improving OCR error correction for Devanagari script languages only. Extending this approach to achieve true multilingual OCR is a complex endeavour. Different languages possess unique linguistic characteristics, script variations, and language-specific nuances. Developing a single model capable of handling this vast diversity effectively remains a challenge. Future work should explore techniques for creating language-agnostic or language-adaptive models to address these limitations and achieve broader multilingual OCR applicability.

\section*{Ethical Statement}
This research utilizes datasets that are openly available in the public domain. The data employed for generating artificial data in this study was sourced from publicly accessible repositories, ensuring no privacy or ethical concerns associated with their use. Specifically, the datasets used do not contain any personally identifiable information or sensitive data that could infringe on individual privacy.

The datasets were chosen based on their availability and openness for research purposes, aligning with ethical guidelines and best practices in data usage. By leveraging publicly available data, this study adheres to the principles of transparency and reproducibility in research while maintaining high ethical standards.

\section*{Acknowledgements}
The authors thank the anonymous reviewers for their constructive feedback and discussion during the rebuttal, which helped improve this submission. We extend our sincere gratitude to the Computation for Indian Language Technology (CFILT) Lab at the Indian Institute of Technology Bombay for providing the computational resources that were indispensable for the successful completion of this research. The first author would like to thank Himanshu Dutta, Sourabh Deoghare, and P S V N Bhavani Shankar for their invaluable support and assistance in conducting the experiments, which contributed to the progress and quality of this work.

\label{sec:bibtex}

\bibliography{acl2023}
\bibliographystyle{acl_natbib}

\newpage
\section{Appendix}

\subsection{PIL (Python Image Library)}
\label{sec:pil}
The Python Imaging Library, commonly known as PIL, is particularly well-suited for image archival and batch-processing applications. 
Pillow\footnote{\url{https://pypi.org/project/pillow}}, an extension of PIL (Python Image Library), stands out as a crucial module for image processing in Python. We generated the images using PIL with dimensions 300x300 and text with a font size of 16.

\subsection{Pytesseract}
\label{sec:pytesseract} Pytesseract\footnote{\url{https://pypi.org/project/pytesseract}} acts as a wrapper around Tesseract OCR engine.
Tesseract\footnote{\url{https://github.com/tesseract-ocr/tesseract}} is an open-source OCR engine designed to extract printed or handwritten text from images. Tesseract boasts support for language recognition in over 100 languages straight out of the box. Since it's open-source, it allows flexibility for customization, integration, and experimentation, which is beneficial in research contexts like error correction. Tesseract is lightweight and can be run on various platforms without requiring extensive computational resources. In contrast, commercial models like Google Vision or OCR engines like Ocular may involve higher resource consumption or come with usage restrictions or costs.

\end{document}